\documentclass[letterpaper, 10 pt, conference]{ieeeconf}  

\IEEEoverridecommandlockouts
\usepackage{textcomp}
\usepackage{xcolor}
\def\BibTeX{{\rm B\kern-.05em{\sc i\kern-.025em b}\kern-.08em
    T\kern-.1667em\lower.7ex\hbox{E}\kern-.125emX}}

\usepackage{algorithm}
\usepackage{algpseudocode}
\algnewcommand\algorithmicinput{\textbf{Input:}}
\algnewcommand\Input{\item[\algorithmicinput]}
\algnewcommand\algorithmicoutput{\textbf{Output:}}
\algnewcommand\Output{\item[\algorithmicoutput]}
\makeatletter
\algnewcommand{\LineComment}[1]{%
  \Statex \hskip\ALG@thistlm 
  \ifnum\theALG@nested>0 \hskip\algorithmicindent\fi 
  // #1%
}
\makeatother

\usepackage{xspace}
\usepackage{cite}
\usepackage{bm}

\usepackage{comment}

\usepackage[toc,acronyms,nopostdot,nogroupskip, nonumberlist]{glossaries-extra} 
\MFUhyphentrue % very important! used to capitalize also word after hyphen in the list of acronyms
\glssetcategoryattribute{general}{glossdesc}{title}
\glssetcategoryattribute{abbreviation}{glossdesc}{title}
\makeatletter
\newglossarystyle{long-initcapsdesc}{%
  \setglossarystyle{long}%
    \renewcommand{\glossentry}[2]{%
    \glsentryitem{##1}\glstarget{##1}{\glossentryname{##1}} &
    \protected@edef\thisdesc{\glsentrydesc{##1}}%
    \xcapitalisewords{\thisdesc}\glspostdescription\space ##2\tabularnewline
  }%
}
\makeatother
\makeglossaries
\setglossarystyle{index}

\newabbreviation[shortplural={POPs}, longplural={polynomial optimization problems}]{POP}{POP}{polynomial optimization problem}
\newabbreviation[shortplural={SDPs}, longplural={semidefinite programs}]{SDP}{SDP}{semidefinite program}
\newabbreviation[shortplural={MILPs}, longplural={mixed integer linear programs}]{MILP}{MILP}{mixed integer linear program}
\newabbreviation[shortplural={CPPs}, longplural={completely positive programs}]{CPP}{CPP}{completely positive program}
\newabbreviation{CP}{CP}{completely positive}
\newabbreviation[shortplural={DNNs}, longplural={doubly non-negative programs}]{DNN}{DNN}{doubly non-negative program}
\newabbreviation[shortplural={LPs}, longplural={linear programs}]{LP}{LP}{linear program}
\newabbreviation[shortplural={SOS}, longplural={sums of squares}]{SOS}{SOS}{sum of squares}
\newabbreviation[shortplural={QCQPs}, longplural={quadratically constrained quadratic programs}]{QCQP}{QCQP}{quadratically constrained quadratic program}
\newabbreviation[]{LICQ}{LICQ}{linear independence constraint qualification}
\newabbreviation[shortplural={NLPs}, longplural={non-linear programs}]{NLP}{NLP}{non-linear program}
\newabbreviation{BM}{BM}{Burer--Monteiro}{}
\newabbreviation{PSD}{PSD}{positive semidefinite}{}
\newabbreviation{LOBPCG}{LOBPCG}{locally optimal block preconditioned conjugate gradient descent}{}
\newabbreviation{ReLU}{ReLU}{rectified linear unit}{}
\newabbreviation{KKT}{KKT}{Karush–Kuhn–Tucker}{}
\newabbreviation[shortplural={NNs}, longplural={neural networks}]{NN}{NN}{neural network}{}
\newabbreviation{FF}{FF}{feed-forward}{}
\newabbreviation{COP}{COP}{copositive}
\newabbreviation{RS}{RS}{Riemannian Staircase}
\newabbreviation{QP}{QP}{quadratic program}

\usepackage[dvipsnames]{xcolor} % ensure this is included

\definecolor{myComment}{rgb}{0.1, 0.1, 1.0} % standard bright blue

\usepackage{graphics} % for pdf, bitmapped graphics files
\usepackage{epsfig} % for postscript graphics files
\usepackage{amsmath} % assumes amsmath package installed
\usepackage{amssymb}  % assumes amsmath package installed
\usepackage{graphicx}
\usepackage{textcomp}

\usepackage{amsfonts}  
\usepackage{mathrsfs}  
\usepackage{float}
\usepackage{lettrine}

\usepackage[utf8]{inputenc}
\usepackage{booktabs}    % For nicer horizontal rules
\usepackage{caption}     % For caption styling (optional)
\usepackage{dblfloatfix}  % improves two-column float placement
\usepackage{placeins}     % provides \FloatBarrier
\usepackage{wrapfig,lipsum,booktabs}
\usepackage{float}
\usepackage{placeins}  % if you still need \FloatBarrier
\usepackage{balance}
\usepackage{afterpage}

\usepackage{wasysym} % Import wasysym

\usepackage{subcaption}
\usepackage{tabularx} % Add to preamble

\newtheorem{prop}{Proposition}

\newcommand{\cp}[1]{\ifmmode {\mathcal{#1}}\else ${\mathcal{#1}}$\fi}
\usepackage{array}
\newcolumntype{P}[1]{>{\centering\arraybackslash}p{#1}}

\usepackage{nccmath}

\def\credrev2{\textcolor{red}}
\def\credrev{\textcolor{red}}

\definecolor{darkgreen}{rgb}{0., 0.4, 0.}
\definecolor{amber}{rgb}{1.0, 0.49, 0.0}
\definecolor{orange}{rgb}{1.0, 0.4, 0.0}

\def \transpose{^\mathsf{T}}
\DeclareMathOperator{\rank}{rank}

\def \Cone{\mathcal{K}}

\def \R{\mathbb{R}}
\def \Sym{\mathbb{S}}
\def \PSD{\Sym_{+}}
\def \CP{\mathbb{C}_*}
\def \COP{\mathbb{C}}
\def \NN{\mathbb{R}_+}
\def \DNN{\PSD^\optvars \cap \NN^\optvars}

\def \opt{^\star}

\newcommand{\constraintindeq}{i}
\newcommand{\numconstraintseq}{m_1}

\newcommand{\constraintindineq}{j}
\newcommand{\numconstraintsineq}{m_2}

\newcommand{\numnetworklayers}{\ell}

\newcommand{\layerdim}{h}

\newcommand{\neuron}{z}

\newcommand{\optvars}{n}
\newcommand{\probdim}{r}
\newcommand{\rankrestriction}{p}

\newcommand{\lagrangemultineq}{\gamma}

\newcommand{\cert}{S}
\newcommand{\cost}{Q}

\usepackage{hyperref}
\hypersetup{colorlinks=true,allcolors=blue}
\makeatletter
\let\NAT@parse\undefined
\makeatother
\begin{document}

\title{$(\text{DNN})^2$: Doubly Non-Negative Relaxations for Deep Neural Networks
%Certifying Optimality of Doubly Non-Negative Relaxations for Neural Network Verification
%A New Perspective on Neural Network Verification using Completely Positive Programming
%Tight Certification of Adversarially Trained Neural Networks using Nonconvex Low-Rank Completely Positive Programming and Semidefinite Relaxations\\
\thanks{%The authors are with the Robust Autonomy and the Autonomy \& Intelligence Labs, Institute for Experiential Robotics, Northeastern University, 360 Huntington Ave, Boston, MA 02115, USA. \texttt{\{zhang.hanna, m.everett, d.rosen\}@northeastern.edu} and \texttt{apapalia@umich.edu}. 
Corresponding author: \texttt{zhang.hanna@northeastern.edu}. We acknowledge the support of the Natural Sciences and Engineering Research Council of Canada (NSERC), the Northeastern University Institute for Experiential Robotics Postdoctoral Fellowship, and Army Research Lab awards W911NF-24-2-006 and W911NF-24-2-0017.}
}
% \author{\IEEEauthorblockN{Hanna Jiamei Zhang, Alan Hilby-Papalia, Michael Everett, and David M. Rosen}
% \IEEEauthorblockA{\textit{Department of Electrical and Computer Engineering} \\
% \textit{Northeastern University}\\
% Boston, MA, USA \\
% \{zhang.hanna, a.hilby-papalia, m.everett, d.rosen\}@northeastern.edu}
% }
% \author{\IEEEauthorblockN{Hanna Jiamei Zhang$^{1}$, Alan Papalia$^{2}$, Michael Everett$^{1}$, and David M. Rosen$^{1}$}
% \IEEEauthorblockA{$^{1}$\textit{Department of Computer Science, Electrical and Computer Engineering, and Mathematics
% Northeastern University, Boston, MA, USA}\\
% \{zhang.hanna, m.everett, d.rosen\}@northeastern.edu}
% \IEEEauthorblockA{$^{2}$\textit{Department of Naval Architecture and Marine Engineering,
% University of Michigan, Ann Arbor, MI, USA}\\
% papalia@umich.edu}
% }
\author{
  Hanna Jiamei Zhang$^{1}$%
  \thanks{$^{1}$Departments of Computer Science, Electrical and Computer Engineering, and Mathematics Northeastern University, Boston, MA.}%
  %\thanks{$^{2}$Department of Electrical and Computer Engineering, Northeastern University, Boston, MA.}%
  %\thanks{$^{3}$Department of Mathematics, Northeastern University, Boston, MA.}%
  \thanks{$^{2}$Department of Naval Architecture and Marine Engineering, University of Michigan, Ann Arbor, MI.}%
  , Alan Papalia$^{2}$,
  Michael Everett$^{1}$, and David M. Rosen$^{1}$
}

\maketitle

\begin{abstract}
% Past: Others have used low-rank factorizations to tractably large scale SDPs (conic programs) effectively. Structure in the CPPs suggests we can leverage the same mechanism. 

% In this work, we propose to leverage low-rank factorizations to solve completely positive programs. Furthermore, leveraging duality theory we hope to provide certificates of global optimality. 

% We propose a nonconvex certification technique, based on a low-rank restriction of a completely positive programming (CPP). The nonconvex relaxation makes strong certifications of a computationally intractable CPP, while optimizing over dramatically fewer variables comparable to much weaker LP methods. Despite non-convexity, we show how off-the-shelf local optimization algorithms can be used to achieve and to certify global optimality in ? time. 
% \gls{SOS}
%Existing \gls{LP} and \gls{SDP} relaxations for \gls{ReLU} \gls{NN} verification omit constraints critical to faithfully representing the network, leaving significant relaxation gaps and overly-conservative safety guarantees. The \gls{CPP} formulation~\cite{brown2022unified} closes this gap exactly but is NP-hard to solve. Its cheapest tractable relaxation, the \gls{DNN}, retains all
Existing \gls{LP} and \gls{SDP} relaxations for \gls{ReLU} \gls{NN} verification yield overly-conservative safety guarantees due to significant relaxation gaps. While the \gls{CPP} formulation~\cite{brown2022unified} closes this gap, it is NP-hard to solve. Its cheapest tractable relaxation, the \gls{DNN}, retains critical constraints as an \gls{SDP}, but one whose size exceeds the reach of interior-point methods at practical scale. %Its cheapest tractable relaxation, the \gls{DNN}, retains all critical constraints while remaining an \gls{SDP}, tractable in principle, but one for which standard interior-point methods fail to scale. 
While \gls{BM} factorization has been applied to make \gls{SDP}-based verification~\cite{chiu2023tight} scalable, no such result exists for the strictly tighter \gls{DNN} formulation. A key obstacle is that additional non-negativity constraints in the \gls{DNN} cause dual multipliers for optimality certification to be \textit{non-unique}, making standard certification methods inapplicable. We propose a novel eigenvalue maximization procedure that searches the non-unique multiplier space for a valid certificate, i.e. a global optimality guarantee. Experiments demonstrate that our approach $(\text{DNN})^2$ produces bounds consistently tighter than the standard \gls{SDP} method, often matching the exact solution, and that our certification procedure confirms global optimality when a valid certificate exists. These results are a key step toward providing tight, certifiable, and computationally scalable verification guarantees needed to deploy neural network controllers and perception modules in safety-critical autonomous systems.
\end{abstract}
%  even these tighter second-order bounds can fail to verify properties that hold, leaving safety-critical systems without guarantees.

% \begin{IEEEkeywords}
% Burer-Monteiro, Semidefinite Programming, Completely Positive Programming, Neural networks, Uncertain systems, Robotics
% \end{IEEEkeywords}

\section{Introduction}

%Neural networks are super useful but have been shown to fail in surprising and unexpected ways, preventing their broad deployment in safety critical settings... One way to address this is with formal verification of safety rules specified as input-output relationships describing limits on expected behavior of a network. 

As \glspl{NN} are rapidly adopted in safety-critical settings (e.g. autonomous vehicles, surgical robots, aerospace controllers), the absence of behavioral guarantees poses a substantial risk. One way to address this is through \textit{neural network verification} methods, which verify or falsify whether a network's outputs remain within safe limits by computing \textit{outer bounds} on the output over a specified region of possible inputs. These outer bounds over-approximate a network's reachable outputs, connecting verification to the reachability analysis underlying reach-avoid guarantees in learning-based control~\cite{everett2021verifcontrol}. % Such over-approximations underpin reach-avoid guarantees in learning-based control, certifying that a system driven by a learned controller or perception module stays within a safe set and away from unsafe states~\cite{everett2021verifcontrol}. 
We focus on \gls{FF} \gls{ReLU} networks, for which verification can be posed as a nonconvex optimization problem that is NP-hard~\cite{katz2017reluplex}, making convex relaxations a principled path to tractable, sound safety verification. 

The most widely studied relaxations are \gls{LP}-based, strengthened with bound-propagation~\cite{wang2021beta, zhang2018efficient}, and apply broadly across different architectures. These are sound (never incorrectly declare an unsafe network safe) and scalable, but face an inherent barrier to tight verification~\cite{salman2019convex}. Specifically, the \gls{ReLU} activation is characterized by a quadratic complementarity constraint, which \gls{LP} relaxations can only approximate with linear outer bounds. The resulting gap can leave verification queries inconclusive even when the network satisfies the safety property. 
%They cannot capture the non-linearity of \gls{ReLU} activations tightly, producing bounds that may be too loose to meaningfully verify safety properties. 
For modern verifiers, ex. $\alpha,\beta$-CROWN~\cite{wang2021beta}, the default recourse is to recover completeness via branch-and-bound~\cite{tjeng2017evaluating}, subdividing the input space to compensate for the loose \gls{LP} relaxation, effective in many settings, but exponential in the worst case. An alternative relaxation that is already tight needs no branching, but solving it scalably with \gls{BM} factorization requires an extra \textit{solution certification} step which we introduce in this work. Ultimately, \gls{LP} relaxations exchange \textit{tightness} for computational \textit{tractability}. When the gap between a relaxed bound and the true optimum determines whether a safe system can be certified or must default to costly conservative fallbacks, closing it has direct operational value.

\begin{figure}[t!]
    \centering
    \includegraphics[width=0.9\linewidth]{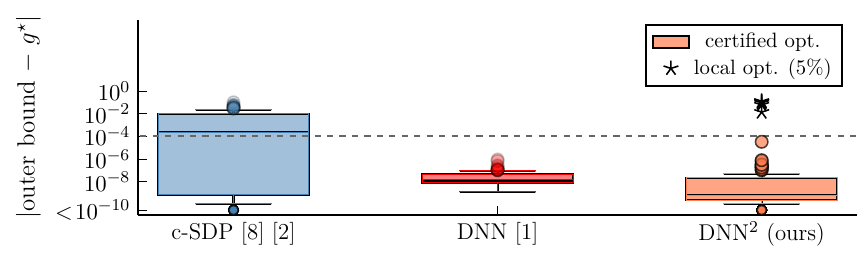}
    \includegraphics[width=0.8\linewidth]{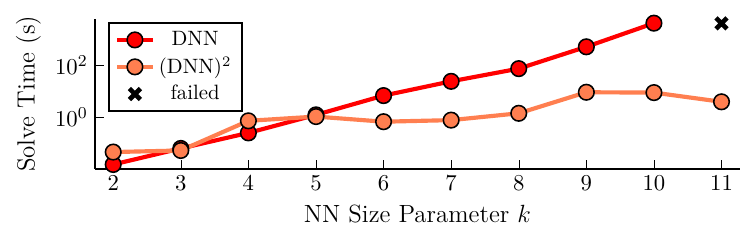}
    \caption{(top) A comparison of the tightness of various verification methods against the ground-truth \gls{MILP} solution across 100 \gls{ReLU} \glspl{NN}. While the standard c-SDP \cite{raghunathan2018semidefinite, chiu2023tight} leaves significant relaxation gaps, our proposed rank-2 BM-factorized DNN, $(\text{DNN})^2$, consistently produces tighter bounds, frequently matching the exact \gls{MILP} solution $95\%$ of all verification instances. Our proposed optimality certification procedure (Eq. \ref{eq:eigenmax_sdp}) reliably distinguishes between global and local optima, with the latter marked with $\mathbf{\APLstar}$. (bottom) Wall time computational performance comparison between the direct \gls{DNN} (solved via interior-point methods, \texttt{MOSEK}) and $(\text{DNN})^2$ (solved as a non-linear program, \texttt{KNITRO}) with increasing network size parameter $k$, using the Julia JuMP interface. As expected, while interior-point solve times grow rapidly (cubically) with problem size, our $(\text{DNN})^2$ exhibits significantly more favorable linear scaling.}
    %Verification bounds relative to ground truth \gls{MILP} solution on 100 \gls{FF} \gls{ReLU} \glspl{NN}. The \gls{DNN} relaxation (solved via rank-2 \gls{BM} factorization) produces bounds consistently \textit{tighter} than the canonical \gls{SDP}\cite{raghunathan2018semidefinite}, and often obtains the exact \gls{MILP} solution. Points determined to be locally optimal $S \not\succeq 0$ with our novel certificate obtained via Eq.~\ref{eq:eigenmax_sdp} show certification failure aligns with instances where the \gls{BM}-\gls{DNN} did not recover the global optimum.}
    %Comparison of \glspl{SDP} (Eq. \ref{eq:gen_cone_program}) and \gls{BM}-factored \glspl{SDP} (Eq. \ref{eq:gen_cone_program_bm}) ($\rankrestriction = 2$) solves with $\Cone=\PSD$ and $\Cone=\DNN$, respectively the canonical \gls{SDP} and \gls{DNN}. We compute the global optimality certificates as \glspl{SDP} (Eq. \ref{eq:eigenmax_sdp}) and mark points that were \textit{not} identified as certifiably globally optimal.}
    \label{fig:dnn_tighter_solve_scale}
\end{figure}

%The most widely studied relaxations are \gls{LP}-based, strengthened with branch-and-bound and bound-propagation techniques~\cite{wang2021beta, zhang2018efficient}. Yet these first-order relaxation approaches face an inherent \textit{barrier} to tight verification~\cite{salman2019convex}, they cannot capture the non-linearity of \gls{ReLU} activations tightly, producing bounds that may be too loose to meaningfully verify safety properties. Ultimately, these relaxations exchange bound \textit{tightness} for computational \textit{tractability}. When the gap between a computed relaxed bound and the true optimum is the difference between certifying a safe system and defaulting to costly conservative fallbacks, closing that gap has direct operational value.
% tightness and shortcomings of current techniques
Beyond \gls{LP} approaches, exact bounds can be obtained with \gls{MILP} verifiers~\cite{tjeng2017evaluating}, but their reliance on exhaustive search renders them intractable for larger problem instances. \gls{SDP} relaxations~\cite{raghunathan2018semidefinite} offer a middle ground between tractability and expressivity. The \gls{PSD} constraint on the lifted moment matrix directly enforces quadratic relationships among neuron variables that would otherwise require an infinite collection of linear constraints to represent. However, existing \glspl{SDP} still omit \textit{verification defining} constraints needed to fully capture the \gls{NN} computation~\cite{brown2022unified}, potentially causing relaxation gaps.

% cpps can overcome the shortcomings?
Brown et al.~\cite{brown2022unified} showed that \gls{FF} \gls{ReLU} verification can be formulated \textit{exactly} (zero relaxation gap) as a \gls{CPP}. Solving a \gls{CPP} is intractable, but unlike the \gls{MILP}, whose intractability offers no intermediate approximations, the \gls{SOS} hierarchy~\cite{parrilo2000structured} provides a family of tractable \gls{CPP} relaxations, each \textit{expressible as an \gls{SDP}}. The coarsest level (0-SOS) yields the \gls{DNN} relaxation, shown on small \glspl{NN}, to be 1) tighter than the \gls{SDP} of~\cite{raghunathan2018semidefinite} and its tightened variants~\cite{batten2021efficient, fazlyab2020safety}, and 2) often exact~\cite{brown2022unified}.  Those results motivate this work, which develops the algorithmic tools needed to exploit these tighter \gls{DNN} formulations at scale.

% scalability issues of sdps solved with BM
Exploiting such formulations requires addressing a second challenge: scalability. Interior-point methods typically used to solve \gls{SDP} relaxations scale as $\mathcal{O}(n^3)$ per iteration in the matrix variable dimension (where $n$ is proportional to the \textit{number of neurons} in the \gls{NN}). \gls{BM} factorization~\cite{burer2003nonlinear} can exploit low-rank structure to solve large-scale \glspl{SDP} at a fraction of this cost. However, this converts the convex relaxation into a \emph{nonconvex} \gls{NLP}: a locally optimal solution does not yield a valid verification bound unless it can be \emph{certified} as a global minimizer of the original \gls{SDP} relaxation. This is distinct from the ``certified'' label commonly used in the \gls{NN} verification literature, which refers to the validity of the relaxation itself. 
%Work in certifiably correct estimation~\cite{rosen2019se, rosen2020Scalable} showed that \gls{BM} factorization of the \gls{SDP} relaxation~\cite{burer2003nonlinear} can exploit low-rank structure to solve large-scale \glspl{SDP} at a fraction of the interior-point cost. However, this converts the convex relaxation into a \emph{nonconvex} problem, introducing local minima: a locally optimal solution does not yield a valid verification bound unless it can be \textit{certified} as a global minimizer of the original relaxation. We note a distinction in terminology, this is not the same as the ``certified'' or ``sound'' label commonly used in the verification literature, which refers to the validity of the relaxation itself. 
Using \gls{BM} factorization within a \gls{RS} framework~\cite{boumal2015riemannian} efficiently yields \textit{certifiably} globally optimal solutions to large-scale \glspl{SDP}, as demonstrated in certifiably correct estimation~\cite{rosen2019se, rosen2020Scalable} and in \gls{NN} verification~\cite{chiu2023tight}, for what we will refer to as the \textit{canonical \gls{SDP}} (c-\gls{SDP})~\cite{raghunathan2018semidefinite}.

Extending these techniques to the tighter \gls{DNN} formulation poses a certification challenge: \gls{KKT} multipliers exist under mild constraint qualifications, but the \gls{LICQ}, which guarantees their \emph{uniqueness}, is generically violated for the \gls{DNN} formulation. Without uniqueness, the \gls{NLP} solver returns one set of multipliers from a family of valid ones, and that particular set may not yield a certificate confirming global optimality of the local \gls{NLP} solution even when it is in fact globally optimal. Certification therefore requires searching over this family, which is the problem we address.

When c-\gls{SDP} bounds are insufficiently tight to verify properties of a \gls{FF} \gls{ReLU} neural network, the tighter \gls{DNN} relaxation formulation can close the gap. To exploit and further investigate the \gls{DNN} relaxation our contributions are:
\begin{itemize}
    \item \textbf{Scalable optimization of the \gls{DNN} relaxation via \gls{BM} factorization.} We apply rank-$2$ \gls{BM} factorization to the \gls{DNN} relaxation, replacing interior-point \gls{SDP} solves with much lower-dimensional \gls{NLP} solves. The resulting bounds are tighter than the c-\gls{SDP} and often exact.
    \item \textbf{Global optimality certificates without the \gls{LICQ}.} We introduce an eigenvalue maximization procedure that searches the set of valid multipliers to construct {global optimality certificates} of the \gls{BM}-\gls{DNN}, a key step toward enabling its practical use in \gls{NN} verification.
\end{itemize}

% \subsection{Related Works}
% \subsubsection{Sound and Complete Verifiers.}
% MILP, SMT...
% they rely on exhaustive search.

% \subsubsection{Sound but Incomplete Verifiers.}
% Bypass exhaustive search by solving convex relaxations of the verification problem.

% \textit{first order relaxations} (opt vars are deg 1 monomials of neuron values): LP, IBP, dual of the relaxed nonconvex problem. Layer-wise convex relaxations exhibits a non-trivial gap. Thus such relaxations cannot capture ReLU, complementarity is naturally represented with second-order (i.e. quadratic) constraints.

% \textit{second order relaxations} (opt vars are deg 2 monomials of neuron values, thus typically require factoring):  Lift to a 2nd order space where ReLU can be linearized. Use RLT to get SDP (Mayhar). SDP from primal perspective. Diagonally dominant matrices to relax SDP more and get an LP.  

% Challenge with second order relaxations: neuron values are obfuscated in the problem formulation...

\textit{Notation.} For a matrix $M$, we use $M_{i,j}$ to denote the entry in the $i$th row and $j$th column, $M_{*, j}$ denotes the entire $j$th column. The matrix/trace inner product is $\langle A, B \rangle = \text{Tr}(A^\top B)$. We write $[n] \triangleq {1, \dots, n}$ for $n > 0$ as a shorthand for sets of indexing integers. $\R$ and $\R_+$ denote the real and nonnegative real numbers, and $\R^n$ and $\R_+^n$ for their $n$-dimensional counterparts. $\Sym^n$ and $\Sym^n_+$ are symmetric and \gls{PSD} matrices of order $n$.  For $x, y \in \R^n$, $x \odot y$ is the Hadamard (elementwise) product. $\mathbf{1}_n \in \mathbb{R}^n$ is the vector of all ones.
%% ============================================================
%% PROBLEM FORMULATION
%% ============================================================
\section{Problem Formulation and Preliminaries}\label{sec:formulation}
We formulate verification for \gls{FF} \gls{ReLU} networks as a nonconvex optimization problem, following the modeling
of~\cite{raghunathan2018semidefinite} and~\cite{brown2022unified}. The restriction to \gls{FF} \gls{ReLU} architectures is significant: the piecewise-linear structure of \gls{ReLU} activations admits an exact representation through bilinear complementarity constraints, which in turn enables the conic reformulations we focus the following discussion on. 

\subsection{\gls{NN} Verification as Optimization}\label{subsec:verif_opt}
We consider an $\numnetworklayers$ layer \gls{FF} \gls{ReLU} \gls{NN} representing function $f$ with input $\neuron_{0, *} \in \mathbb{R}^{\layerdim_0}$ and output $f(\neuron_{0,*})=\neuron_{\numnetworklayers,*} \in \mathbb{R}^{\layerdim_\numnetworklayers}$, with $f$ being the composition of functions $f = f_{\numnetworklayers} \circ f_{\numnetworklayers-1} \circ \cdots \circ f_{1}$. The $i$th layer of $f$ is function $f_i:\R^{\layerdim_{i-1}} \rightarrow \R^{\layerdim_i}$
\begin{equation}\label{eq:network_function}
    \neuron_{i,*} = f_i(\neuron_{i-1,*}) = \sigma(\hat{\neuron}_{i,*}) = \sigma(\overline{W}[i-1]\neuron_{i-1,*}+b[i-1])
\end{equation}
where  $\layerdim_i$ is the dimension of the hidden variable $\neuron_{i,*}$, $\overline{W}[i-1] \in \R^{\layerdim_i \times \layerdim_{i-1}}$ are the weight matrices, and $b[i-1] \in \R^{\layerdim_i}$ are bias vectors. The vector of all neurons in the $i$th layer is denoted by $\neuron_{i, *}$ and $\neuron_{i, j}$ refers to the $j$th neuron in the $i$th layer. The activation function is the \gls{ReLU} $\sigma(\hat{\neuron}_{i,j})=\text{max}(0,\hat{\neuron}_{i,j})$ which acts elementwise on vectors and is not applied to the last layer $f_\numnetworklayers$. Preactivation values are denoted $\hat{z}_{i,*}$ implying that $\neuron_{0, *} = \hat{\neuron}_{0, *}$. Introduced in \cite{brown2022unified}, a positive/negative splitting for each neuron $\lambda^+_{i,j} = \neuron_{i,j}$ and $ \lambda^-_{i,j} = \neuron_{i,j} - \hat{\neuron}_{i,j}$ such that $\lambda^+, \lambda^- \geq 0$ is used rather than the typical pre/post-activation parameterization. %The following variable substitution is performed 
% \begin{equation}
%     \lambda^+_{i,j} = \neuron_{i,j} \quad \lambda^-_{i,j} = \neuron_{i,j} - \hat{\neuron}_{i,j}.\label{eq:neuron_splits}    
% \end{equation}

\textit{Safety rules} are specified as input-output relationships on $f$. For all inputs ${\neuron}_{0, *} = \mathcal{X} \subseteq \R^{\layerdim_0}$, we want to ensure that the output belongs to set ${\neuron}_{\numnetworklayers, *}=\mathcal{Y} \subseteq \R^{\layerdim_\numnetworklayers}$. Following common convention, we consider bounded, polytopic input sets $\mathcal{X}$ and output sets $\mathcal{Y}$ defined by a half-space constraint:
\begin{equation}
    \mathcal{X} = \{x\in\R^{\layerdim_0} \, | \, \overline{A}x \leq a\}, \; \mathcal{Y} = \{y\in\R^{\layerdim_\numnetworklayers} \, | \, \overline{c}^\top y\geq d\}\footnote{%An overline, $\overline{M}$ denotes matrices that act on a single network layer and an unmarked $M$ denotes matrices that act on a collection of \textit{all} positive/negative splittings of neurons 
    Overline notation, $\overline{M}$,  taken from \cite{brown2022unified}.} 
    \label{eq:inout_safe}
\end{equation}

Substituting $\lambda^+, \lambda^-$ %\eqref{eq:neuron_splits} 
into the \gls{QCQP} for \gls{FF} \gls{ReLU} \gls{NN} verification \cite{raghunathan2018semidefinite} yields the non-convex, non-linear \gls{QCQP}:
\begin{equation}\label{eq:verif_nonconvex}
\begin{aligned}
  g^\star = &\min_{\lambda^+,\, \lambda^-} \; c^\top
  (\lambda^+_{\numnetworklayers, *} - \lambda^-_{\numnetworklayers, *})\\
     \text{s.t.} \quad&A(\lambda^+_{0, *} - \lambda^-_{0, *}) \leq a, \\
    &\lambda^+_{i+1, *} - \lambda^-_{i+1, *}
      = W[i]\lambda^+_{i, *} + b[i], \quad\forall i \in [\numnetworklayers]\\
    &\lambda^+_{i,j}\, \lambda^-_{i,j} = 0, \quad \forall \, i \in [\numnetworklayers-1], \, j \in \layerdim_{i}\\
    &\lambda^+,\, \lambda^- \geq 0
\end{aligned}
\end{equation}
where the network is safe if and only if $g^\star \geq d$ as per \eqref{eq:inout_safe}. The optimal value $g^\star$ of~\eqref{eq:verif_nonconvex} is the ground truth against which all relaxation bounds in this work are measured, i.e. a relaxation is \emph{tight} to the extent that its optimal value approaches~$g^\star$.

\subsection{Generalized Cone Programs}\label{subsec:gen_conic_prog}
A generalized cone program (or conic optimization) is a convex optimization problem class that minimizes a linear objective function over the intersection of an affine subspace and a convex cone $\Cone^\optvars$, unifying \glspl{SDP}, \glspl{DNN}, and \glspl{CPP} as
\begin{equation}\label{eq:gen_cone_program}\tag{Cn}
     g_{\Cone}\opt = \min_{Z \in \Cone^\optvars} \; g(Z) \; \text{ s.t. } \mathcal{A}(Z)=b, \; \mathcal{B}(Z) \leq u
\end{equation}
where $Z \in \Sym^\optvars$ is a symmetric matrix variable %\footnote{For notational convenience, we describe the generalized cone program for only symmetric cones $\Cone \in \Sym^\optvars$, excluding \glspl{LP}, SOCPs, etc.}
%excluding \glspl{LP} with $\Cone = \R^\optvars$ and $Z\in \R^\optvars$.} 
constrained to lie in a cone of symmetric matrices $\Cone^\optvars \subset \Sym^\optvars$, $f: \Sym^\optvars \rightarrow \R$ is a convex and twice-continuously-differentiable function, and $\mathcal{A}: \Sym^\optvars \rightarrow \R^{\numconstraintseq}$, and $\mathcal{B}: \Sym^\optvars \rightarrow\R^{\numconstraintsineq}$ are linear operators defined by $\mathcal{A}(Z)_\constraintindeq = \left\langle A_\constraintindeq, Z\right\rangle, \constraintindeq\in[\numconstraintseq]$ and $ 
    \mathcal{B}(Z)_\constraintindineq = \left\langle A_\constraintindineq, Z\right\rangle, \constraintindineq\in[\numconstraintsineq].$ The choice of $\Cone^\optvars$ influences both the tightness and tractability of a given~\eqref{eq:gen_cone_program}. We discuss these choices of convex $\Cone^\optvars$: the \gls{PSD} cone $\PSD^\optvars$ and the cone of \gls{CP} matrices $\CP^\optvars$, that is, matrices that have a factorization with entry-wise non-negative entries:
\begin{align}
    \PSD^\optvars := &\{X \in \Sym^\optvars \,|\, v^\top Xv \geq 0, \forall v \in \R^\optvars\}, \\ 
    \text{ or equivalently } &\{X \in \Sym^\optvars \,|\, X = BB^\top, B \in \R^{\optvars\times \probdim}\}\\
    \CP^\optvars := &\{X \in \Sym^\optvars \,|\, X = BB^\top, B \in \R^{\optvars\times \probdim}_{+}\}.
\end{align}
% \item and the cone of \gls{COP} matrices
% \begin{equation}
%     \COP^n := \{X \in \Sym^n  | v^\topXv \geq 0, \forall v \in \R^n_{\geq 0}\}.
% \end{equation}
The nomenclature of \gls{SDP} and \gls{CPP} arises naturally for describing problems of the form~\eqref{eq:gen_cone_program} with the corresponding choice of $\Cone^\optvars$. % The $\CP$ and $\COP$ cones are duals of each other~\cite{dur2010copositive} and are both proper convex cones.
Unlike the \gls{PSD} cone, merely testing membership in $\CP^\optvars$ is NP-hard~\cite{dur2010copositive}, making direct optimization over it intractable. This obstacle motivates the use of tractable outer approximations, one being the intersection of the \gls{PSD} and entrywise non-negative cones $\PSD^\optvars \cap \NN^\optvars$, the \gls{DNN} cone. The relationship between these cones is 

%In contrast to the $\PSD$ matrix cone, the $\CP$ and $\COP$ matrix cones are not self-dual, as the notation suggests they are in fact duals of eachother \cite{dur2010copositive}. They are both proper and convex cones, and thus inherit the straightforward analysis of convex optimization, which we leverage in this work to provide efficient solution recovery and global optimality guarantees. As they are not self-dual (i.e. have self concordant barrier functions), interior point methods that work well for enforcing $\Cone^\optvars=\PSD$ in \glspl{SDP} are insufficient for enforcing $\Cone^\optvars=\CP$ in \glspl{CPP} and $\Cone^\optvars = \COP$ \glspl{COP}. Methods for doing so remains an active area of research. 

% \begin{align*}
%     \CP^n  \subseteq \PSD^n \cap \NN_n  \subseteq \quad \PSD^n \quad \subseteq \PSD^n + \NN^n 
%     \subseteq \COP^n
% \end{align*}

\begin{equation}\label{eq:relaxation_hierarchy}
  \underbrace{\CP^\optvars}_{\text{\gls{CPP} (exact)}}
  \;\subseteq\;
  \underbrace{\PSD^\optvars \cap \NN^\optvars}_{\text{\gls{DNN} (0-SOS)}}
  \;\subseteq\;
  \underbrace{\PSD^\optvars}_{\text{\gls{SDP}}}.
  %\;\subseteq\;   
  %\underbrace{\PSD + \NN}   
  %{\PSD + \NN}   
  %\;\subseteq\;   
  %\underbrace{\COP}_{\text{\gls{COP}}}
\end{equation}

\subsection{Conic Reformulations of Neural Network Verification}\label{subsec:conic_reform}
%In the remainder of this section, we present the standard \gls{SDP} relaxation~\cite{raghunathan2018semidefinite}, the \gls{DNN} (0-\gls{SOS}) relaxation of the \gls{CPP}, and the exact \gls{CPP}~\cite{brown2022unified} forms of \gls{NN} verification as instances of a \textit{generalized cone program} to highlight both their relationship to one another and the gaps this work addresses in how to approach scalably solving each form.
We now present three conic reformulations of~\eqref{eq:verif_nonconvex}: 1. the canonical \gls{SDP} relaxation \cite{raghunathan2018semidefinite}, 2. the exact reformulation of~\eqref{eq:verif_nonconvex} as a \gls{CPP}~\cite{brown2022unified}, and 3. its tractable relaxations (in particular the \gls{DNN}). We emphasize that each arises from a different construction but shares the common form of a generalized cone program~\eqref{eq:gen_cone_program} to highlight both their relationship to one another and the gaps this work addresses in how to approach scalably solving each form. Note that the structural properties outlined for each instance of \eqref{eq:gen_cone_program} extend beyond the \gls{NN} verification setting, in fact they hold more generally for the entire $\Cone^\optvars$ problem class.

As the objective and constraint functions are all \textit{quadratic}, written more generally \eqref{eq:verif_nonconvex} can be expressed as
\begin{equation}
\label{eq:qcqp_def}\tag{Qcqp}
    \begin{aligned}
    g_{\text{QCQP}}\opt =  \min_{X \in \R^{\optvars \times \probdim}} & \left\langle \cost, X X\transpose\right\rangle \\
    \text { s.t. } & \left\langle A_\constraintindeq, X X\transpose\right\rangle = b_\constraintindeq, \quad \constraintindeq\in[\numconstraintseq]\\
    & \left\langle A_\constraintindineq, X X\transpose\right\rangle \leq u_\constraintindineq, \quad \constraintindineq\in[\numconstraintsineq], 
    \end{aligned}
\end{equation}
with objective matrix $Q \in \Sym^\optvars$, constraint matrices $A_\constraintindeq, A_\constraintindineq\in \Sym^\optvars$, constraint vectors $b\in \R^{\numconstraintseq} \; u \in\R^{\numconstraintsineq}$, and $\probdim \le \optvars$.  Since $X \in \R^{\optvars \times \probdim}$ only enters \eqref{eq:qcqp_def} through the outer product $XX\transpose$, which is symmetric and \gls{PSD} satisfying $\rank(XX\transpose) \le \probdim$, convex relaxations can be obtained by replacing $XX^\top$ with matrix variable $Z$ constrained to an appropriate cone $\Cone^\optvars$.

\subsubsection{SDP relaxation ($\Cone^\optvars = \PSD^\optvars$)} The class of problems of the form \eqref{eq:qcqp_def} admits a general procedure called Shor's relaxation \cite{shor1987quadratic} for constructing convex relaxations. This consists of replacing the low-rank symmetric outer product $X X\transpose$ in \eqref{eq:qcqp_def} with a \textit{generic} symmetric and \gls{PSD} matrix variable $Z \in \Cone^\optvars = \PSD^\optvars$ in \eqref{eq:gen_cone_program}, resulting in the canonical \gls{SDP} relaxation of~\eqref{eq:verif_nonconvex} we denote $g_{\PSD}$, first proposed in \cite{raghunathan2018semidefinite}.

Shor's relaxation is solvable in polynomial time via interior-point methods and provides a lower bound $g_{\PSD}\opt \leq g_{\text{Qcqp}}\opt$. However, it increases the dimension of the decision variable from $X \in \R^{\optvars \times \probdim}$ to $Z \in \PSD^{\optvars}$, so that the number of free variables scales quadratically in network size, limiting applicability due to memory constraints. This concern persists for the \gls{DNN}. More fundamentally, the \gls{PSD} constraint alone does not enforce entry-wise non-negativity of $Z$, which~\cite{brown2022unified} showed to be among the most critical verification-defining constraints: in ablation experiments, omitting $\Lambda \geq 0$ produced relative errors ranging from $10^2$ to $10^4$.

% \textcolor{red}{PSD cone easy to solve}
% \textcolor{blue}{EXPAND on commentary about this problem form pros/cons using Certi-FGO stuff: Better expressive space with second order terms, but larger one (way more variables, $\optvars \times \probdim$ to $\optvars^2$).}

\subsubsection{CPP form ($\Cone^\optvars = \CP^\optvars$)} Analogous to Shor's relaxation for \glspl{SDP}, Brown et al.~\cite{brown2022unified} constructed an \emph{exact} convex formulation of~\eqref{eq:verif_nonconvex} by replacing $XX^\top$ where $X=\bigl(\lambda \; 1 \bigr)$ with matrix variable $Z = \bigl(\begin{smallmatrix} \Lambda & \lambda \\ \lambda^\top & 1 \end{smallmatrix}\bigr)$ constrained to be \gls{CP} ($Z \in \CP^\optvars$)~\cite[Thm.~4.1]{brown2022unified}. $\Lambda$ is a matrix which represents the second-order cross terms of elements in $\lambda = \bigl((\lambda^+)^\top\; (\lambda^-)^\top\; s^\top\bigr)^\top$ and $s$  is the concatenated slack variables used to represent inequality constraints. They form a minimal set of \emph{verification-defining} constraints whose removal fundamentally misrepresents the \gls{NN}~\cite[Eqns. 13b-g]{brown2022unified} which we restate as \eqref{eq:verif_nonconvex} being \textit{equivalent} to \eqref{eq:gen_cone_program} with $\Cone^\optvars = \CP^\optvars$, a \gls{CPP} we denote $g_{\CP}$. 

Despite being convex, optimizing over $\CP^\optvars$ is intractable as membership testing in the $\CP^\optvars$ cone and its dual, the copositive cone $\COP^\optvars$, is NP-hard and co-NP-complete respectively~\cite{dur2010copositive}.  The intractability of the \gls{CPP} echoes that of the \gls{MILP}, both provide exact formulations of~\eqref{eq:verif_nonconvex}, but package the NP-hardness differently.  Whereas the \gls{MILP} encodes it in combinatorial branching, the \gls{CPP} encodes it entirely in the cone constraint, simplifying the objective and feasibility constraints to be linear~\cite{burer_copositive_2009}.  This structural property enables the hierarchy of tractable relaxations described next. 

% \textcolor{red}{CPP cone hard to solve}

% \textcolor{blue}{
% Despite being convex, co-positive programs are generally intractable because checking checking membership in the co-positive cone $A \in \mathcal{CP}$ is co-NP-complete. Intuitively, checking membership in the dual, i.e. complete positivity $A \in \mathcal{C}$ should have the same complexity, and indeed it does as shown in \cite{dickinson_computational_2014}. Both weak and strong membership tests for both these cones are NP-hard. The choice to focus on co-positive optimization problems is non-restrictive. In a landmark result, Burer demonstrated an exact mapping from mixed-binary quadratic problems (MBQP) to linear optimization problems over the cone of completely positive matrices $\mathcal{CP}$—dualizing these problems yields a linear optimization problem with a single co-positivity $\COP$ constraint \cite{burer_copositive_2009}. This suggests that the complexity (non-convexity) of the solution is shifted to the cone membership constraint and away from the cost/feasible set which are linear with respect to the cone in question.}

\subsubsection{DNN relaxation ($\Cone^\optvars = \PSD^\optvars \cap \NN^\optvars$)}
Although optimizing over $\CP^\optvars$ is intractable, Parrilo~\cite{parrilo2000structured} constructed a hierarchy of tractable outer approximations of $\CP^\optvars$, expressible as \glspl{SDP}, with successively tighter approximations at greater computational cost. The coarsest ($0$-\gls{SOS}) level of this hierarchy relaxes the \gls{CPP} by replacing $\CP^\optvars$ with $\Cone^\optvars = \PSD^\optvars \cap \NN^\optvars$, a \gls{DNN} we denote $g_{\PSD \cap \NN}$. For brevity, we refer to this as the \gls{DNN} relaxation, noting that it is an \gls{SDP} because it is an instance of \eqref{eq:gen_cone_program} with $\Cone^\optvars=\PSD^\optvars$. Unlike the c-SDP relaxation of \eqref{eq:verif_nonconvex}, the \gls{DNN} retains all verification-defining constraints and is thus tighter, $g_{\text{Qcqp}}\opt = g_{\CP}\opt \geq g_{\PSD \cap  \NN}\opt \geq g_{\PSD}\opt$~\cite{brown2022unified}.
%Whether these inequalities are strict is an empirical question we investigate in the experiments.
%Unlike the canonical \gls{SDP}, which relaxes~\eqref{eq:verif_nonconvex} directly, the \gls{DNN} relaxes the exact \gls{CPP} formulation and retains all verification-defining constraints.  It is therefore strictly tighter: $g_{\text{Qcqp}}\opt = g_{\CP^\optvars}\opt \geq g_{\Sym_+ \cap  \NN}\opt \geq g_{\Sym_+}\opt$.

The \gls{DNN} represents a potentially favorable tradeoff between tightness and tractability. It is the tightest relaxation that remains expressible as an \gls{SDP} of the same size as the \gls{CPP} formulation, without requiring additional decision variables or constraints as in higher levels of the SOS hierarchy. It retains all verification-defining constraints identified in \cite{brown2022unified}, and empirically achieves significantly smaller relaxation gaps than existing SDP formulations for small networks. While even tighter SDP relaxations may be possible using higher SOS levels, the 0-SOS relaxation is a compelling target for scalable solver development and investigation of the tightness-tractability frontier. Thus, we focus on the DNN relaxation as the tightest practically available formulation, with the exact CPP serving as a theoretical reference point.% The following section develops the Burer-Monteiro (BM) optimization framework and the certification machinery required to tackle the DNN relaxation.
\subsection{Scalably Solving SDPs via Burer–Monteiro Factorization}\label{subsec:scalably_solve_sdps}
%For $\Cone = \PSD$ and $\Cone = \Sym_+ \cap \NN$, the conic program~\eqref{eq:gen_cone_program} is an \gls{SDP} that can in principle be solved directly via interior-point methods. However, these methods scale as $\mathcal{O}(n^3)$ per iteration and require storage and manipulation of the dense symmetric matrix decision variable $Z\in\Sym^n$ which scales $\mathcal{O}(n^2)$, making solving large instances prohibitive. 
\gls{BM} factorization of \glspl{SDP} dramatically reduces the number of decision variables from $\optvars^2$ to $\optvars \probdim$. When the relaxation is exact, the minimizer $Z\opt$ admits a low-rank factorization $Z\opt = X\opt {X\opt}\transpose$ with $X\opt \in \R^{\optvars \times \probdim}$, a global minimizer of~\eqref{eq:qcqp_def}.  Burer and Monteiro (\gls{BM})~\cite{burer2003nonlinear} proposed to algorithmically exploit this observation by reparameterizing $Z$ using an \textit{assumed} rank-$\rankrestriction$ factorization $Z = YY\transpose$ with \textit{factor} $Y \in \R^{\optvars \times \rankrestriction}$, $\probdim \leq \rankrestriction \ll \optvars$. Substituting this parameterization into~\eqref{eq:gen_cone_program} yields the \gls{BM} factorization, which enforces positive semi-definiteness by construction for $\mathcal{K} = \PSD^\optvars$.
\begin{equation}
\label{eq:gen_cone_program_bm}\tag{Bm}
    g_{BM}\opt = \min_{Y \in \R^{\optvars \times \rankrestriction}} \; g(YY^\top) \; \text{ s.t. } \; \mathcal{A}(YY^\top)=b, \;\mathcal{B}(YY^\top) \leq u.
\end{equation}

\subsection{Global Optimality Certification and the Role of the LICQ}
Solving these problem forms requires addressing that the \gls{BM} factorized form \eqref{eq:gen_cone_program_bm} is a \textit{non-convex} \gls{NLP} that is known to admit suboptimal local minima. To overcome this, the \gls{RS} framework has been employed successfully as a systematic approach to search for global optimum by incrementally increasing the rank of the factor matrix $Y$, thereby smoothing the optimization landscape and escaping sub-optimal points. Within such a framework, we must check that first-order stationary points $Y\opt$ of the \gls{NLP}~\eqref{eq:gen_cone_program_bm} with corresponding Lagrange multipliers $\mu = (\lambda, \gamma)$, actually correspond to a solution $Z \in \PSD^\optvars$ of~\eqref{eq:gen_cone_program}, our target \gls{SDP} relaxation. That is, specifically we are checking the optimality of $Z=Y\opt {Y\opt}^\top$ as a solution to the \gls{SDP} using the \textit{certificate matrix}
\begin{equation}\label{eq:cert_matrix}
  S = S(Y\opt, \mu) \triangleq \nabla f(Y\opt {Y\opt}\transpose)
  - \mathcal{A}^*(\lambda) - \mathcal{B}^*(\gamma),
\end{equation}
where $\mathcal{A}^*: \R^{\numconstraintseq} \rightarrow \Sym^\optvars $, and $\mathcal{B}^*: \R^{\numconstraintsineq} \rightarrow \Sym^\optvars$ linear operators defined as $\mathcal{A}^*(\lambda) = \sum^{\numconstraintseq}_{i=1}\lambda_iA_i$ and $\mathcal{B}^*(\gamma) = \sum^{\numconstraintsineq}_{j=1}\gamma_j A_j$.
% The \gls{RS} framework can be effectively applied to \glspl{SDP} for which the \textit{\gls{LICQ} holds}.

The \gls{RS}~\cite{boumal2015riemannian} integrates these components~\cite[Algo. 1]{rosen2020Scalable}: starting from a rank-$\rankrestriction$ factorization, it iteratively performs local optimization to recover a \gls{KKT} point $Y\opt$ of~\eqref{eq:gen_cone_program_bm}, constructs the certificate matrix $S$, either certifying optimality of $Z=Y\opt {Y\opt}^\top$ when $S \succeq 0$, or escaping to rank-$(\rankrestriction\!+\!1)$ otherwise~\cite[Thm. 4]{rosen2020Scalable}. Prior applications of \gls{RS} relied on the \gls{LICQ} holding at the local minimizer $Y\opt$, guaranteeing the existence and uniqueness of the multipliers $\mu$ to construct $S$ for optimality verification and establish a second-order direction of descent when $Y\opt$ is found not to be optimal. %The correctness of this framework relies on the \gls{LICQ} holding at $Y\opt$, which guarantees the existence of a unique set of Lagrange multipliers $\mu$ implying that if $S \not\succeq 0$, non-optimality of $Y\opt$ and a second-order direction of descent are definitively established. 
In certifiable perception, the \gls{LICQ} is known to hold generically for the constraints arising in many robotic perception tasks (which often involve estimation over smooth manifolds), allowing the successful application of this framework~\cite{rosen2019se, papalia2025overviewburermonteiromethodcertifiable}. Similarly, this framework has been applied in \gls{NN} verification for the canonical \gls{SDP}~\cite{raghunathan2018semidefinite}, because of \cite[Lemma E.2]{chiu2023tight}, if the nonzero preactivation (NPCQ) condition holds, then the \gls{LICQ} holds. We refer the reader to~\cite{papalia2025overviewburermonteiromethodcertifiable} for a detailed treatment of the role of \gls{LICQ} in a \gls{RS} framework.

\section{Certifiably and Scalably Solving DNNs}
Here we present our approach to scalably solving \gls{DNN} relaxations of the form~\eqref{eq:gen_cone_program} via \gls{BM} factorization, followed by a certification procedure that provides global optimality guarantees even when the \gls{LICQ} fails.
%We describe an algorithmic framework for scalably and efficiently solving the~\eqref{eq:gen_cone_program} programs for \gls{CP} and \gls{DNN} cones that are \textit{contained within} the \gls{PSD} cone described in Section~\ref{subsec:conic_reform} using \gls{BM} factorization and the \gls{RS} meta-algorithm. Here we present our main technical contributions: extending this framework to the \gls{DNN} relaxation and the exact CPP with novel optimality certification procedures. In particular, our approach is an attempt to generalize to cases where the \gls{LICQ} condition does not hold, a limitation of prior work in this area.

%In this section, we describe an algorithmic approach for solving large-scale instances of Shor-relaxed conic programs \eqref{eq:gen_cone_program_shor_relaxed} \textcolor{red}{efficiently?}.  This is a non-linear program and solving is subject to local optima. In the following section, we discuss methods on how to identify when a returned locally optimal solution is in fact the global optima, as well as a framework for using local solves to efficiently extract the global optimum. 

% and the novel certification machinery needed to provide global optimality guarantees for \gls{DNN} and \gls{CPP} forms. 

%% --- A: BM + Staircase (established framework) ----------------
\subsection{Burer--Monteiro Factorization for DNN Scalablility}\label{sec:bm_staircase}
%\textcolor{red}{We make the observation that for $\Cone = \CP$, the addition of linear constraint $Y \geq 0$ (element-wise positivity of the low-rank matrix variable) suffices to enforce \gls{CP}. THIS NEEDS A NEW PROB FORMULATION BC IT'S CONSTRAINTS ACTING ON FACTORS!} Furthermore, for $\Cone = \DNN$, the addition of constraints enforcing element-wise non-negativity on the individual entries of the resultant outer product $(YY\transpose)_{ij} \geq 0 \; \forall i,j \in [\optvars]$ are required to enforce double non-negativity. We now have \gls{BM} factorizations, i.e. \glspl{NLP} forms, of the target \gls{SDP}, \gls{CPP}, and \gls{DNN} at our disposal.
The \gls{DNN}, \eqref{eq:gen_cone_program} with $\Cone^\optvars = \DNN$, can be written in the form of \eqref{eq:gen_cone_program_bm} with the addition of constraints enforcing element-wise non-negativity on individual entries of the resultant outer product $(YY\transpose)_{ij} \geq 0 \; \forall i,j \in [\optvars]$. This is the \gls{BM} factorization of the target \gls{DNN}, \gls{BM}-\gls{DNN}.

\begin{prop}\label{rm:certification}
Let $Y\opt$ be a first-order stationary point of~\eqref{eq:gen_cone_program_bm}. If there exist dual multipliers $\mu = (\lambda, \gamma)$ satisfying the \gls{KKT} conditions at $Y\opt$ such that $S \succeq 0$ as per~\eqref{eq:cert_matrix}, then $Z\opt =Y\opt{Y\opt}\transpose$ is a global minimizer of~\eqref{eq:gen_cone_program}.
%Suppose~\eqref{eq:gen_cone_program} satisfies Slater's constraint qualification, and let $Y\opt$ be a first-order stationary point of~\eqref{eq:gen_cone_program_bm}. If there exist dual multipliers $\mu = (\lambda, \gamma)$ satisfying the \gls{KKT} conditions at $Y\opt$ such that $S \succeq 0$, then $Z\opt = Y\opt {Y\opt}\transpose$ is a global minimizer of~\eqref{eq:gen_cone_program}.
\end{prop}

The optimality certificate described in Prop. \ref{rm:certification} requires only the \emph{existence} of multipliers $\mu$ such that $\cert \succeq 0$, \textit{regardless} of the \gls{LICQ}. With the \gls{LICQ}, $\mu$ is \textit{unique} \cite{wachsmuth2013licq}, allowing for efficient certification (a linear system solve to form $\cert$ directly)~\cite{papalia2025overviewburermonteiromethodcertifiable} and that $\cert \not\succeq 0$ definitively proves non-optimality of $Y\opt$\cite[Footnote~3]{rosen2020Scalable}. This case has been explored extensively in past work (Sec. \ref{subsec:scalably_solve_sdps}). Without the \gls{LICQ}, there exists more than one set of valid multipliers $\mu$, so $\cert \not\succeq 0$, with certificate $S$ constructed from the \textit{particular} multipliers $\mu\opt$ returned by the specific \gls{NLP} solver used, is \textit{inconclusive}. Certification reverts to a semidefinite feasibility problem, finding an intersection between the affine space and the \gls{PSD} cone for a certificate (if it exists), which is generally as computationally demanding as the original \gls{SDP} \eqref{eq:gen_cone_program} \cite{papalia2025overviewburermonteiromethodcertifiable}.

The \gls{LICQ} is violated for \glspl{DNN} due to the simultaneously active entry-wise non-negativity constraints. A unique $\mu$ does not exist and we require a robust procedure to search over the space of feasible multipliers of \gls{BM}-\gls{DNN} for a $\mu$ satisfying $S\succeq 0$, which, if found, certifies global optimality of solution $Z=Y\opt{Y\opt}^\top$ for the \gls{DNN} as per Prop.~\ref{rm:certification}.

\subsection{Search for Optimality Certificate}\label{sec:eigenmax}
We formalize the search for a certificate as the following optimization problem, in which we maximize the minimum eigenvalue of the certificate matrix $S$ over the space of feasible multipliers $\mu$ of the \gls{BM}-\gls{DNN}:
\begin{equation}\label{eq:eigenmax}
  \max_{\mu=\lambda, \gamma} \,
 \text{eigmin} \circ \cert
  \;\; \text{s.t.} \;
  \begin{aligned}
    &\cert\, Y\opt = 0, \; c\odot \gamma = 0, \; \gamma \leq 0.
  \end{aligned}
\end{equation}
$Y\opt$ is a stationary point returned from the \gls{NLP} solver, the constraint slack is $c=\bigl(\mathcal{B}(Y\opt {Y\opt}\transpose) - u\bigr)$, and $\text{eigmin}$ returns the minimum eigenvalue of a given matrix. The constraints of \eqref{eq:eigenmax} enforce \gls{KKT} conditions of \gls{BM}-\gls{DNN} (stationarity, complementary slackness, and dual feasibility) in order for $\mu$ to represent true stationary points of \eqref{eq:gen_cone_program_bm}. Prob. ~\eqref{eq:eigenmax} is convex but nonsmooth. If the optimal value, $\theta_{\min}^\star \ge 0,$ then a multiplier vector $\mu$ such that $S \succeq 0$ exists, Prop.~\ref{rm:certification} applies, and the \gls{BM} solution is certified globally optimal. If it is negative, and the \gls{DNN} satisfies a suitable constraint qualification (CQ) (ex. Slater's), then $Z=Y\opt(Y\opt)T^\top$ is \emph{not} optimal for the \gls{DNN}. This is due to the fact for convex problems, the \gls{KKT} conditions are necessary and sufficient for global optimality under a suitable CQ, so the non-existence of valid multipliers is conclusive. 
%The remaining challenge in this case is translating this non-optimality of $Z\opt$ in the \gls{DNN} into a second-order direction of descent for $Y$ in the \gls{BM}-factorized problem~\eqref{eq:gen_cone_program_bm}, which is a distinct nonconvex problem with a different feasible set. Developing such a descent procedure is the subject of future work; in our experiments, all tested instances were successfully certified at the first \gls{BM} solve without requiring rank escalation.%If it is negative, the certification result is \textit{inconclusive}: the solution $Y\opt$ may still be optimal, but this cannot be established from the given $\mu\opt$. 
%\textcolor{red}{$Y\opt$ may or may not be the global optimum but there is no way to determine.} % (over an exhaustive search) and the $\mu$ at termination provides the descent direction for rank escalation as per Thm.~\ref{thm:certification}(b).}

In practice, \gls{NLP} solvers satisfy the \gls{KKT} conditions only to within solver tolerances, so the exact stationarity and complementary slackness constraints in \eqref{eq:eigenmax} may admit no feasible point in finite precision. To ensure the feasible set has interior, we relax these constraints using $\ell_1$-norm bounds $\epsilon_{\text{stat}}$ and $\epsilon_{\text{cs}}$, set from the residuals of the multipliers $\mu\opt$ by the \gls{NLP} solver. We note that one could select different norms that would interact differently with the solver of choice. Thus, let $\cp{F}$ denote the feasible set $\cp{F} = \left\{ \mu \mid \|\cert Y \opt \|_1 \leq \epsilon_{\text{stat}}, \; \| c\odot \lagrangemultineq \|_1 \leq \epsilon_{\text{cs}}, \; \lagrangemultineq \leq 0 \right\}$.

To express the objective of \eqref{eq:eigenmax} operationally, we introduce an epigraph variable $\theta_\text{min}$ representing the minimum eigenvalue of $S$ yielding the numerically robust \gls{SDP} 
\begin{equation}\label{eq:eigenmax_sdp}
  \max_{\mu,\, \theta_\text{min}} \;\; \theta_\text{min}
  \quad \text{s.t.} \quad
  \begin{aligned}
    &\cert(\mu)- \theta_\text{min}\, I \succeq 0, \quad \mu \in \cp{F},
  \end{aligned}
\end{equation}
%which unfortunately has the same dimensionality and solve difficulty of the original \gls{DNN} relaxation. For the purposes of this work, to demonstrate the framework we solve this problem as an \gls{SDP} and leave more scalable solves of this for future work.
where results $\theta_\text{min}\opt \geq 0$ means $S \succeq 0$ and $\theta_\text{min}\opt < 0$ means $S \not\succeq 0$. %In its presented form,~\eqref{eq:eigenmax_sdp} is an \gls{SDP} of the same dimension as the original \gls{DNN} relaxation. Our scalability contribution addresses \emph{solving} the  \gls{DNN}~\eqref{eq:gen_cone_program} via \gls{BM} factorization rather than interior point methods; the separate certificate cost reflects a validation choice, not an intrinsic property of the certificate. Indeed,~\eqref{eq:eigenmax_sdp} need not be solved to global optimality: any feasible $\mu$ such that $\cert(\mu)\succeq 0$ suffices. For demonstration our certificate is solved as an \gls{SDP} using interior-point methods here, deferring more scalable strategies to future work.
Solving~\eqref{eq:eigenmax_sdp} as an \gls{SDP} (of the same dimension as our target \gls{DNN}) establishes that a valid certificate \textit{can} be constructed despite the obstacle of \gls{LICQ} failure. NB: In this work, solving~\eqref{eq:eigenmax_sdp} via interior-point methods is a validation choice, \textit{not} an intrinsic cost of the certificate, since \textit{any} feasible $\mu$ yielding $\cert \succeq 0$ suffices, ~\eqref{eq:eigenmax_sdp} need not be solved to global optimality, permitting early termination. Scaling the certificate search~\eqref{eq:eigenmax} is the remaining step toward an end-to-end pipeline, which we leave to future work.% Scalably conducting the certificate search~\eqref{eq:eigenmax} is the remaining step toward a fully scalable pipeline, which we leave to future work.

\section{Experimental Results and Discussion}
We consider \gls{FF} \gls{ReLU} networks, generated by selecting weights and biases uniformly between $-1.0$ and $1.0$. We consider safety rule $\mathcal{X} := \{\neuron_{0,*}|\neuron_{0,*}\in [-1.0, 0.1]^2 \}$  and $\mathcal{Y} = \{\neuron_{\numnetworklayers,*}|\neuron_{\numnetworklayers,*}\geq 0\}$. We use notation $h=[2, 10, 1]$ to represent the \gls{NN} from Brown~\cite{brown2022unified} with two inputs, one hidden layer with 10 neurons, and 1 output. We computed ground-truth bounds by solving the \gls{MILP}~\cite{tjeng2017evaluating}. \glspl{SDP} are solved using \texttt{MOSEK}~\cite{mosek} and \gls{BM}-\glspl{SDP} (i.e. \glspl{NLP}) with \texttt{KNITRO}~\cite{byrd2006knitro}. All solvers are set with comparable tolerances. \glspl{NLP} are initialized with the network state obtained by evaluating the network at the midpoint of the input set. Our implementation is available as open-source Julia/JuMP code.\footnote{\url{https://github.com/hanjzh/dnn-sq}} We evaluate the rank $\rankrestriction=2$ \gls{BM} factorization of the \gls{DNN} relaxation, denoted $(\text{DNN})^2$, to demonstrate both the tightness gains of the \textit{cheapest} \gls{BM}-\gls{DNN} over the c-\gls{SDP} and the reliability of our global optimality certificate~\eqref{eq:eigenmax_sdp}. We restrict attention to these instances to isolate the effect of relaxation tightness and applying \gls{BM} factorization.%; a thorough comparison against complete search-based verifiers is left to future work.

\subsection{Global Optimality Certification and Tightness of DNN}
On the small test \gls{NN} of~\cite{brown2022unified}, where the \gls{DNN} relaxation is known to be tighter than c-\gls{SDP} and often exact~\cite{brown2022unified}, we find that $(\text{DNN})^2$ recovers the (exact) global optimum of~\eqref{eq:gen_cone_program} in 95\% of verification queries, and that our certificate~\eqref{eq:eigenmax_sdp} correctly identifies each such case (Fig.~\ref{fig:dnn_tighter_solve_scale}). %This suggests that the \gls{BM}-factorized \gls{DNN} is a practical pathway to solving the verification problem to global optimality.

\begin{figure}
    \centering
    \begin{subfigure}{0.9\linewidth}
        \includegraphics[width=\linewidth]{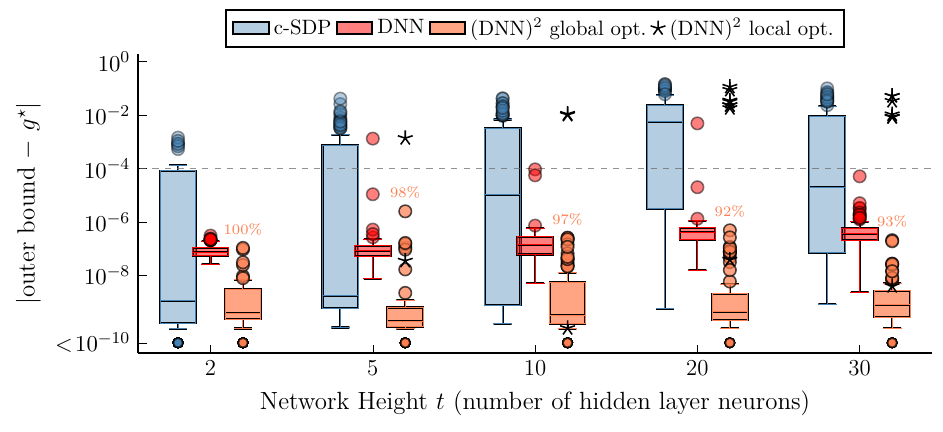}
        % \caption{Changing height of network.}
    \end{subfigure}
    \begin{subfigure}{0.9\linewidth}
        \includegraphics[width=\linewidth]{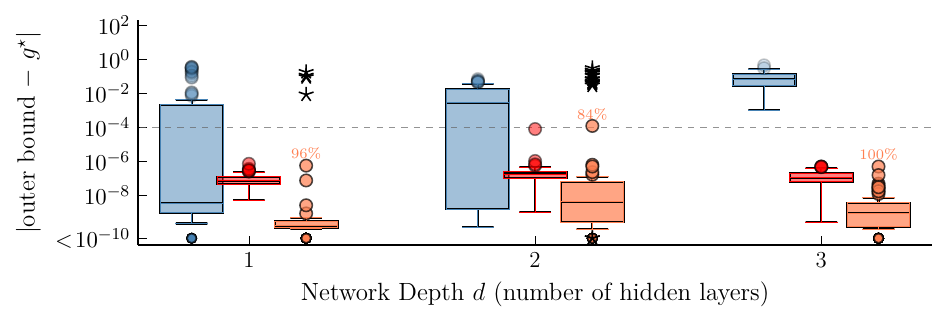}
        % \caption{Changing depth of network.}
    \end{subfigure}
    \hfill
    \caption{Analysis of how verification accuracy degrades as \glspl{NN} increase in height (top) and depth (bottom). The canonical c-\gls{SDP} bounds (blue) become progressively looser as the network grows, making safety verification more difficult. In contrast, our $(\text{DNN})^2$ formulation (orange) remains highly tight to the MILP ground truth across all tested scales, demonstrating its robustness for larger, more complex network architectures. Our proposed certification procedure (Eq. \ref{eq:eigenmax_sdp}) successfully identifies instances where $(\text{DNN})^2$ has reached a \textit{local} rather than \textit{global} optimum; these points are marked with $\mathbf{\APLstar}$ and are excluded from the box plots. The percentage of instances certified as globally optimal is annotated for each configuration.}
    %Verification bounds scalably obtained using c-\gls{SDP} and $\rankrestriction=2$ \gls{BM}-\gls{DNN} problem formulations for increasing network height (top) and depth (bottom). Our proposed certificate \eqref{eq:eigenmax_sdp} is solved using an interior point method (MOSEK), with points identified as local omitted from the BM-DNN $(\rankrestriction=2)$ boxes and marked explicitly with $\star$. For the BM-DNN $(\rankrestriction=2)$ our certificate identified local optima in percentages 5.0\%, 2.5\%, 2.0\%, 17.6\%, 23.1\% for heights $h=2, 5, 10, 20, 30$ and 2.9\%, 12.7\%, 12.6\% for depths $d=1, 2, 3$.}
    \label{fig:network_comparison}
\end{figure}

%Fig. \ref{fig:dnn_tighter} is presented for $h=[2, 10, 1]$ recreates finding of \cite{brown2022unified} in that the \gls{DNN} is tighter than the \gls{SDP}. What we find, solving the \gls{BM}-\gls{DNN} level $\rankrestriction=2$ rank-restriction is that it often produces that global optimum, and that we can identify that 95\% of the time. 
% \subsection{Tightness of DNN Relaxation}
To further investigate the practicality of this restriction we compare the c-\gls{SDP}, the \gls{DNN}, and our $(\text{DNN})^2$ varying \gls{NN} height $t$, and depth with number of hidden layers $d$, evaluating 100 random network instances per configuration for architectures $h=[2, t, 1]$ and $h = [2, 4 \cdot \mathbf{1}_{d}, 2]$. %$h = [2, \underbrace{2, \ldots, 2}_{d}, 2]$ 
As shown in Fig.~\ref{fig:network_comparison}, the \gls{SDP} relaxation bounds become progressively looser for larger (wider and deeper) networks, whereas $(\text{DNN})^2$ remains tight (i.e. within meaningful tolerance $10^{-4}$ marked in all figures with a dotted grey line) to the \gls{MILP} ground truth in the majority of instances. Our certificate correctly identifies solutions that are \emph{not} globally optimal, i.e., it never falsely certifies a suboptimal solution. 

In $0.075\%$ of queries, the $(\text{DNN})^2$ bound matches the \gls{MILP} to numerical precision yet our certificate fails to confirm global optimality. This reflects a solver limitation, not a flaw in the formulation: the interior-point method used for~\eqref{eq:eigenmax_sdp} times out before finding a $\mu$ with $S \succeq 0$. Since any such $\mu$ suffices (global optimality of~\eqref{eq:eigenmax_sdp} is not required) more scalable solution strategies are a promising direction for future work. Developing these would enable certified \gls{DNN} verification at the same scale as the \gls{BM} solve itself. 

Finally, we note that the \gls{DNN} relaxation is exact in all but 2 instances, and the $(\text{DNN})^2$ recovers the globally optimal \gls{DNN} bound in $84$--$100\%$ of queries depending on network configuration. Developing a \gls{RS}-style rank escalation procedure for the \gls{DNN} setting (Section~\ref{sec:eigenmax}) would, with appropriate theoretical foundations, guarantee recovery of the global \gls{DNN} optimum in all cases. Whether the \gls{DNN} relaxation remains frequently exact for larger, trained networks is an important open question for future investigation.

%This is not a failure of the certificate formulation itself, but of the interior point method solver used to solve~\eqref{eq:eigenmax_sdp}. The \gls{SDP} solver times out and returns a \textit{feasible} $\mu$, but not one for which $S(\mu) \succeq 0$. A more efficient, scalable solver for~\eqref{eq:eigenmax_sdp} would close this gap.

\begin{figure}[htbp]
    \centering
    \includegraphics[width=0.2\textwidth]{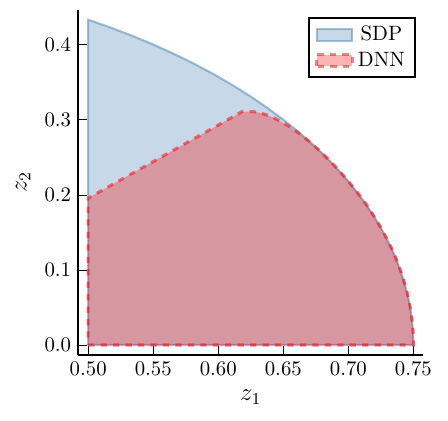}
    \includegraphics[width=0.2\textwidth]{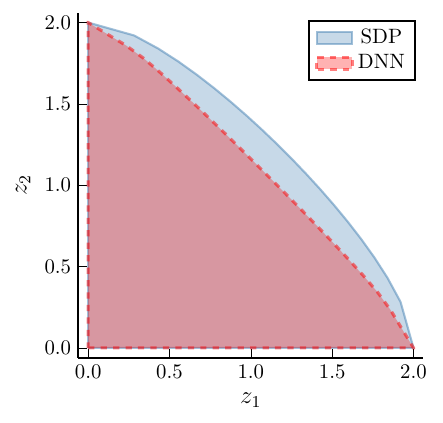}
    % \begin{subfigure}[b]{0.23\textwidth}
    %     \centering
    %     \includegraphics[width=\textwidth]{figures/sdp_vs_dnn_feasible_sets_2b.pdf}
    %     \label{subfig:feas_region_fixed_input}
    % \end{subfigure}
    % \hfill
    % \begin{subfigure}[b]{0.23\textwidth}
    %     \centering
    %     \includegraphics[width=\textwidth]{figures/sdp_vs_dnn_feasible_sets_2c.pdf}
    %     \label{subfig:feas_region_overall}
    % \end{subfigure}
    \caption{%\ref{subfig:single_relu} Visualization of the \gls{LP}, \gls{SDP}, and \gls{DNN} for a single ReLU unit with input x and output z.  
    Visualizing why the \gls{DNN} relaxation is tighter than c-\gls{SDP}. Let $z_1 = \text{ReLU} \left( x_1 + x_2\right)$ and $z_2 = \text{ReLU}\left(x_1 - x_2\right)$. (left) The feasible set of each relaxation projected onto the post-activation variables $(z_1, z_2)$ for fixed input $x_1 = x_2 = 0.5$. (right) The same projection across all feasible inputs, under the objective $\max\; z_1 + z_2$.
    %(left) Shows the feasible region of relaxed activations for a fixed input, both equal to 0.5. (right) Shows the overall feasible region when maximizing the objective $z_1 + z_2$ across all inputs. 
    Because the \gls{DNN} formulation enforces additional element-wise non-negativity constraints not present in the \gls{SDP}, it produces a strictly smaller feasible set (red) than the SDP (blue), leading to tighter outer \gls{NN} bounds and more accurate safety guarantees.}
    \label{fig:dnn_tighter_ragunathan}
\end{figure}

To understand why the \gls{DNN} relaxation is tighter, we replicate the feasible set analysis of \cite[Fig.~2]{raghunathan2018semidefinite} in Fig.~\ref{fig:dnn_tighter_ragunathan}. Both the \gls{SDP} and \gls{DNN} enforce joint constraints that yield a convex feasible region, but the \gls{DNN} additionally imposes element-wise non-negativity, producing a strictly smaller feasible set and thus a strictly tighter relaxation.

\subsection{Computational Scalability of BM-DNN over DNN}
To further motivate this approach, we compare the computational scaling of $(\text{DNN})^2$ solved as an \gls{NLP} against the direct \gls{DNN} solved via interior-point methods. We use a toy network (weights set to $1$ and biases $0$) with a single-point (trivial) input set and architecture $[k,\, k \cdot \mathbf{1}_{k},\, k]$ for increasing $k$. As shown in Fig.~\ref{fig:dnn_tighter_solve_scale}, \gls{DNN} solve times grow rapidly with problem size, while the \gls{BM}-factored form exhibits significantly more favorable scaling. For small instances, the $(\text{DNN})^2$ solve is slower than \gls{DNN} due to \gls{NLP} solver overhead, however, the scaling trends diverge rapidly with increasing $k$. This gap will widen as interior-point costs grow cubically in the matrix dimension while the \gls{BM} variable count grows only linearly. The failed \gls{SDP} solve at $k=11$ due to memory exhaustion illustrates this precisely.
%Moreover, since the \gls{BM} factorization reduces the problem to an \gls{NLP} whose variable count scales linearly rather than quadratically in the network size, this approach lays the groundwork for verification at scales far beyond the reach of interior-point \gls{SDP} solvers. that can be solved scalably to global optimality with \gls{RS}~\cite{chiu2023tight}

\section{Discussion and Future Work}

We have shown that the \gls{DNN}, the coarsest level of the \gls{CPP} relaxation hierarchy, consistently yields tighter verification bounds than the canonical \gls{SDP} and is frequently exact, even when solved as a rank-$2$ \gls{BM} factorization. Our certificate~\eqref{eq:eigenmax_sdp} reliably identifies globally optimal \gls{BM}-\gls{DNN} solutions despite \gls{LICQ} failure. Together these results establish \gls{BM}-factorized \gls{DNN} relaxations as a promising path toward scalable, tight neural network verification.

% Two directions for future work remain. First, \emph{scalable
% certificate computation}: the certificate
% problem~\eqref{eq:eigenmax_sdp} reduces to semidefinite
% \emph{feasibility}---any $\mu \in \cp{F}$ achieving
% $\theta_{\min} \geq 0$ suffices---admitting first-order and
% spectral bundle methods that avoid interior-point solves entirely.
% Second, \emph{rank escalation for the \gls{DNN}}: when certification
% fails, establishing non-optimality of the current iterate and
% extracting a descent direction to drive rank escalation toward the
% global optimum, extending the \gls{RS} framework to this setting.

% \section*{Acknowledgment}
% We acknowledge the support of the Natural Sciences and Engineering Research Council of Canada (NSERC), the Northeastern University Institute for Experiential Robotics Postdoctoral Fellowship, and Army Research Lab awards W911NF-24-2-006 and W911NF-24-2-0017. % Nous remercions le Conseil de recherches en sciences naturelles et en génie du Canada (CRSNG) de son soutien.
\bibliographystyle{IEEEtran}
\bibliography{ref}

\end{document}